\documentclass[9pt,conference]{IEEEtran}
\IEEEoverridecommandlockouts

\usepackage{cite}
\usepackage{amsmath,amssymb,amsfonts}
\usepackage{algorithmic}
\usepackage{graphicx}
\usepackage{textcomp}
\usepackage[x11names]{xcolor}

\usepackage{array}
\usepackage{mathtools}
\usepackage{tcolorbox}
\usepackage{color}
\usepackage{theorem}
\usepackage{amssymb}
\usepackage{caption}
\usepackage{subcaption}
\usepackage{cite,hyperref}
\usepackage{cases}
\usepackage{url}
\usepackage{algorithmic}
\usepackage{algorithm}
\usepackage{enumitem}
\usepackage{multirow}
\usepackage{hhline}
\input{my_styles.sty}

\usepackage{tikz}
\usetikzlibrary{shapes,arrows}
\usepackage{moresize}
\usepackage{pgfplots}
\usepackage{wrapfig}
\pgfplotsset{compat=1.17}
\pgfplotstableset{col sep=comma}
\def \Cbal {\hat{\bbC}_\mathrm{bal}}
\def \Cdeb {\hat{\bbC}_\mathrm{deb}}

\newtcolorbox{myblockt}[1]{colback=urblue!5!white,
	colframe=urblue,fonttitle=\bfseries,
	title=#1}
\newtcolorbox{myblock}{colback=urblue!5!white,
	colframe=urblue,fonttitle=\bfseries}

\def\BibTeX{{\rm B\kern-.05em{\sc i\kern-.025em b}\kern-.08em
    T\kern-.1667em\lower.7ex\hbox{E}\kern-.125emX}}

\def \precite {{\color[RGB]{51,187,238}[ref.]}}
\newcommand \draft[1]        {{\color[RGB]{182,133,103}#1}}

\pgfplotsset{compat=1.17}
\pgfplotstableset{col sep=comma}
\tikzset{every mark/.append style={scale=1.5, solid}, font=\footnotesize}
\pgfplotsset{
    width=1.05\textwidth,
    legend style={
        font=\scriptsize ,  
        inner xsep=1pt,
        inner ysep=1pt,
        nodes={inner sep=1pt}},
    legend cell align=left,
	every axis/.append style={line width=0.5pt},
	every axis plot/.append style={line width=.9pt},
    every axis y label/.append style={yshift=-3pt}
}

\def \fvnnsmp {Red2}
\def \fvnnbal {white!60!Red2}
\def \fvnndeb {black!60!Red2}

\def \lpcasmp {white!0!RoyalBlue2}
\def \lpcabal {white!60!RoyalBlue2}
\def \lpcadeb {black!60!RoyalBlue2}

\def \kpcasmp {DarkGoldenrod3}
\def \kpcabal {white!60!DarkGoldenrod3}
\def \kpcadeb {black!60!DarkGoldenrod3}

\begin{document}

\title{Fair CoVariance Neural Networks
\thanks{This work was partially supported by the NSF under award CCF-2340481. 
Research was sponsored by the Army Research Office and was accomplished under Grant Number W911NF-17-S-0002. The views and conclusions contained in this document are those of the authors and should  not be interpreted as representing the official policies, either expressed or implied, of the Army Research Office or the U.S. Army or the U.S. Government. The U.S. Government is authorized to reproduce and distribute reprints for Government purposes notwithstanding any copyright notation herein. Part of this work was supported by the TU Delft AI Labs programme, the NWO OTP GraSPA proposal \#19497, and the NWO VENI proposal 222.032.
        Emails:  
        \href{mailto:a.cavallo@tudelft.nl}{a.cavallo@tudelft.nl}, 
        \href{mailto:nav@rice.edu}{nav@rice.edu}, 
        \href{mailto:segarra@rice.edu}{segarra@rice.edu},
        \href{mailto:e.isufi-1@tudelft.nl}{e.isufi-1@tudelft.nl}  }
}

\author{
\IEEEauthorblockN{
Andrea Cavallo\IEEEauthorrefmark{1},
Madeline Navarro\IEEEauthorrefmark{2},
Santiago Segarra\IEEEauthorrefmark{2},
Elvin Isufi\IEEEauthorrefmark{1}
} %
\IEEEauthorblockA{
\IEEEauthorrefmark{1}Delft University of Technology, Delft, Netherlands } %
\IEEEauthorblockA{
\IEEEauthorrefmark{2}Rice University, Houston, TX, USA } %
}

\maketitle

\begin{abstract}
    Covariance-based data processing is widespread across signal processing and machine learning applications due to its ability to model data interconnectivities and dependencies. 
    However, harmful biases in the data may become encoded in the sample covariance matrix and cause data-driven methods to treat different subpopulations unfairly.
    Existing works such as fair principal component analysis (PCA) mitigate these effects, but remain unstable in low sample regimes, which in turn may jeopardize the fairness goal.
    To address both biases and instability, we propose Fair coVariance Neural Networks (FVNNs), which perform graph convolutions on the covariance matrix for both fair and accurate predictions.
    Our FVNNs provide a flexible model compatible with several existing bias mitigation techniques.    
    In particular, FVNNs allow for mitigating the bias in two ways: first, they operate on fair covariance estimates that remove biases from their principal components; second, they are trained in an end-to-end fashion via a fairness regularizer in the loss function so that the model parameters are tailored to solve the task directly in a fair manner.
    We prove that FVNNs are intrinsically fairer than analogous PCA approaches thanks to their stability in low sample regimes.
    We validate the robustness and fairness of our model on synthetic and real-world data, showcasing the flexibility of FVNNs along with the tradeoff between fair and accurate performance.
\end{abstract}


\section{Introduction}
\label{S:intro}

Covariance-based learning has a long-standing history as an approach to conveniently model critical information about observed data, boasting success in several applications ranging from brain connectivity estimation~\cite{bessadok2022graph,QIAO2016399} to blind source separation~\cite{Belouchrani1997bss,nikunen2014bss} and financial data analysis~\cite{cardoso2020algorithms,wang2022network}. 
For example, the covariance matrix is the foundation of principal component analysis (PCA)~\cite{Jolliffe2002pca}, the prevailing approach for summarizing high-dimensional data via dimension reduction.
PCA exploits the eigenvectors of the covariance matrix, termed principal components (PCs), which denote primary directions of spatially distributed data.
Beyond PCA, the theoretical and empirical advantages of graph neural networks (GNNs)~\cite{isufi2024graph,gama2020graphs,gao2023trade} have led to the development of covariance neural networks (VNNs), where the covariance matrix is seen as the input graph for a GNN~\cite{sihag2022covariance}.
By spectral graph theory, VNNs can be viewed as an extension of PCA with learnable weights assigned to PCs~\cite[Theorem 1]{sihag2022covariance}; they are transferable across datasets~\cite{sihag2024transferability} and effective in temporal and sparse settings~\cite{cavallo2024stvnn, cavallo2024sparsecovarianceneuralnetworks} and for applications to brain data~\cite{sihag2024explainable,sihag2024towards,sihag2025explainablebrainagegap}.
Moreover, VNNs are provably stable to covariance estimation errors in low sample regimes~\cite[Theorem 2]{sihag2022covariance}, while PCA-based data processing may encounter unexpected behavior if the estimated PCs differ greatly from the true ones~\cite{Jolliffe2002pca}.

These tools have shown great success in extracting rich information from correlated data.
However, real-world data often contains harmful biases such as poor representation of certain communities, resulting in disparate treatment across subpopulations~\cite{chouldechova2017fair}.
If relevant information is correlated with sensitive attributes, then PCs can encode these biases.
Such PCs may yield inaccurate representations or increased discrepancies in treatment of different groups.
For example, segregation may worsen if observed data experiences shifts in distribution across groups~\cite{olfat2019convex}.
Additionally, representations of underrepresented subpopulations may be far more inaccurate, resulting in overdependence on majority groups~\cite{samadi2018price,olfat2019convex}.

Fair learning methods promote unbiased treatment for data-driven tools, with recent interest emerging for fairness in all steps of data processing pipelines, from data representation to predictions~\cite{navarro2024data,zemel2013learning,kamiran2012data}.
While fairness for predictions is the most well-studied task~\cite{mehrabi2022SurveyBiasFairness,caton2024fairnessmachinelearning}, attention of late has turned towards unbiased representation learning~\cite{liu2022fair}, as several real-world datasets exhibit preferential treatment due to unequal representation of different groups~\cite{samadi2018price}.
As the pervasiveness of high-dimensional data increases, recent works attempt dimensionality reduction while mitigating biases in data~\cite{samadi2018price,olfat2019convex,kose2023fairnessawaredimensionality}.
Fair variants of PCA have shown success in reducing biases in projected data, where the goal is either to (i) obtain group-agnostic projections for fair downstream tasks~\cite{olfat2019convex,tan2020learning,kose2023fairnessawaredimensionality} or (ii) to encourage equitable representation accuracy across groups~\cite{samadi2018price,tantipongpipat2019multicriteria,kamani2022efficient,pelegrina2024novelfairpca}. 
Such methods are often inefficient to compute or return suboptimal solutions~\cite{pelegrina2022analysis,lee2022fast,Kleindessner2023efficient,xu2024efficient}.
Additionally, the sensitivity of PCA to outliers and insufficient data remains a challenge since fair PCA approaches are still unstable to minor perturbations.

\medskip 

\noindent \textbf{Contributions.} Motivated by the value of covariance-based learning and the challenge of removing biases in PCs, we propose \textit{fair VNNs (FVNNs)} to exploit the advantages of VNNs for covariance-based learning in unfair settings.
Exploiting the flexibility of VNNs, we introduce fairness by (i) estimating a fairer version of the covariance matrix and (ii) penalizing biases in the training loss.
Given an unbiased covariance estimate, we prove that the natural stability of VNNs leads to fairer outcomes than fair PCA when groups follow different distributions.
Moreover, tuning the weight of the loss penalty allows for flexible control of the tradeoff between fairness and accuracy, whereas PCA is performed separately from any downstream task.
We summarize our contributions as follows.
\begin{itemize}[left= 5pt .. 15pt, noitemsep]
    \item[(i)] 
    We present FVNNs for fair covariance-based model predictions with reduced influence from biased correlations.
    Our model uses fair covariance matrices from transformed data while explicitly promoting unbiased predictions in end-to-end learning.
    \item[(ii)] 
    We theoretically show that the inherent stability of VNNs promotes equitable treatment of different groups.
    \item[(iii)] 
    We empirically validate the stability and flexibility of FVNNs on one synthetic and three real datasets with known biases on both classification and regression tasks. 
\end{itemize}

\section{Problem Statement}

Consider a dataset $\ccalD = \{(\bbx_i,y_i,z_i)\}_{i=1}^T$ of $T$ tuples, each consisting of features $\bbx_i \in \reals^N$, a target $y_i \in \ccalY$, and a group label $z_i \in \{1,\dots,G\}$ for every $i\in \{1,\dots,T\}$.
Depending on the task at hand, $\ccalY$ can be a set of discrete class labels or real-valued regression targets.
Each group $g\in\{1,\dots,G\}$ is associated with a random vector $\boldsymbol{x}^{(g)}$ with mean $\bbmu_g = \mbE[\boldsymbol{x}^{(g)}]$ and covariance matrix $\bbC_g = \mbE[(\boldsymbol{x}^{(g)}-\bbmu_g)(\boldsymbol{x}^{(g)}-\bbmu_g)^\top]$.
If $z_i = g$, then sample $i$ belongs to group $g$, and the feature vector $\bbx_i$ is an instantiation of $\boldsymbol{x}^{(g)}$.
Furthermore, let $\bbZ \in \{0,1\}^{T \times G}$ be the indicator matrix denoting group membership, where $Z_{ig} = 1$ if and only if $z_i = g$.

Our goal is to learn a mapping $\Phi:\reals^N \rightarrow \ccalY$ using covariance information to predict targets $\bby = \{y_i\}_{i=1}^T$ from features $\bbX = [\bbx_1,\dots,\bbx_T]^\top \in \reals^{T\times N}$ such that prediction performance is not biased with respect to group membership $\bbz = \{z_i\}_{i=1}^T$.
With the observed samples, we estimate the mean $\hbmu = \frac{1}{T} \bbX^\top \bbone_{T}$ and covariance matrix $\hbC = \frac{1}{T} (\bbX - \bbone_T \hbmu^\top)^\top(\bbX - \bbone_T \hbmu^\top)$.
Analogously, we define a data matrix for each group $\bbX_g$ containing the $T_g$ samples in group $g$ with the corresponding sample mean $\hbmu_g$ and covariance $\hbC_g$.
Finally, we let $\bby_g$ collect the targets corresponding to group $g$.

We view features as nodes in a graph whose connectivity is described by the covariance matrix $\bbC$, data samples $\bbX$ as graph signals, and we let our model $\Phi$ be a VNN, which performs graph convolutions on the signals~\cite{sihag2022covariance}, followed by a readout layer.
More formally, a VNN architecture stacks $L$ VNN layers, each consisting of a graph convolutional covariance filter bank of size $F_{\text{in}} \times F_{\text{out}}$ followed by a point-wise nonlinearity $\sigma$. 
The covariance filter and the propagation rule for each layer $l = 1, \ldots, L$ and parallel filter $f = 1, \ldots, F_{\text{out}}$ are defined as
\alna{
    \bbH(\bbC) = \sum_{k=0}^Kh_k\bbC^k~~\textnormal{and}~~\bbx^l_f = \sigma\left(\sum_{j = 1}^{F_{\text{in}}}\bbH^l(\bbC)\bbx^{l-1}_j\right),
    \nonumber
}
where $\{h_k\}_{k=0}^K$ denotes the set of learnable filter coefficients.
Let $\Phi(\bbx,\hbC,\ccalH)$ be the VNN architecture followed by a readout layer for the downstream task, where $\bbx$ is the input feature vector, $\hbC$ the input covariance matrix, and $\ccalH$ collects the filter coefficients for all layers.

We aim to learn parameters $\ccalH$ from $\ccalD$ to yield predictions $\hat{y} = \Phi(\bbx, \hbC, \ccalH)$ that are fair with respect to the group label $z$.
While equitable outcomes are desirable, balancing treatment of different groups is also a necessity~\cite{samadi2018price}.
Indeed, the downside of popular fairness definitions such as demographic parity (DP)~\cite{feldman2015certifying} and equality of odds (EO)~\cite{hardt2016equality} is that they focus solely on outcomes without considering if the model exhibits preferential learning for certain groups~\cite{baer2019fairness,kleinberg2016inherent}.
Inaccurate predictions for an underrepresented group can lead the model to consider those samples irrelevant for subsequent training.
In this case, we may satisfy DP or EO at the cost of neglecting certain subpopulations.
Thus, we emphasize equitable attention in training, a goal well-suited to VNNs, which are robust to insufficient data.

We formalize imbalanced treatment across groups as the difference in prediction performance between each pair of groups,
\begin{equation}\label{eq:imbgrp}
    \Delta\ccalL(\bbX,\bby,\bbz) :=
    \sum_{g=1}^G \sum_{h>g}
    \left|
        \ccalL( \bbX_g, \bby_g, \Phi ) - \ccalL( \bbX_h, \bby_h, \Phi )
    \right|,
\end{equation}
where $\mathcal{L}(\bbX_{g}, \bby_{g}, \Phi)$ denotes the loss function measuring the performance of model $\Phi$ on the data samples of group $g$.
Not only is $\Delta\ccalL$ in~\eqref{eq:imbgrp} analogous to the goal of equal reconstruction error for fair PCA, but it aligns with other notions of fairness such as bounded group loss~\cite{agarwal2019fair}.

\section{Methodology}
\label{s:method}

We promote fairness for FVNN predictions in two ways.
First, as with some PCA-based approaches~\cite{pelegrina2024novelfairpca,kose2023fairnessawaredimensionality}, we consider a fair version of the sample covariance matrix using fair data preprocessing techniques.
Second, since VNNs can be trained end-to-end for a downstream task, we explicitly encourage unbiased predictions by penalizing biases in the loss function to be minimized, allowing control of the trade-off between fairness and accuracy during training.

\subsection{Fair covariance matrices}
\label{ss:faircov}

FVNNs are a general framework that can accommodate any fair covariance estimation technique. We exemplify them with two in particular that promote different goals.
First, for $G=2$ with one group poorly represented, we may consider a balanced covariance matrix estimate~\cite{pelegrina2024novelfairpca}
\alna{
    \Cbal
    &~=~&
    \alpha \hbC
    +
    (1-\alpha) (\hat{\mathbf{C}}_h - \hat{\mathbf{C}}_g) = \alpha_g\hbC_g + \alpha_h\hbC_h,
\label{eq::cbal}
}
where $\hbC_h$ is the sample covariance of the disadvantaged group and $\hbC_g$ the other, $\alpha \in [0,1]$ is the balancing term, $\alpha_g = (\alpha T_g/T + \alpha - 1)$ and $\alpha_h = (\alpha T_h/T + 1 - \alpha)$.
The estimate $\Cbal$ was defined for fair PCA to yield equal reconstruction error across groups~\cite{pelegrina2022analysis} as it interpolates between the original sample covariance $\hbC$ and the attempt to reduce the discrepancy between the minority and majority covariances $\hbC_h$ and $\hbC_g$, respectively.
Alternatively, we may wish to remove the dependence on groups in the covariance matrix to avoid predictions that are based on sensitive attributes.
To this end, for $\beta\in[0,1]$ we present
\alna{
    \Cdeb
    &~=~&
    \bbX^\top ( \bbI_T + \beta \bbZ\bbZ^\top )^{-1} \bbX / T,
\label{eq::cdeb}
}
which essentially computes the sample covariance matrix for transformed data $(\bbI_T + \beta \bbZ\bbZ^\top)^{-1/2}\bbX$.
While $\Cdeb$ was originally proposed for group-agnostic PCA projections~\cite{kose2023fairnessawaredimensionality}, we consider the transformed covariance $\Cdeb$ for FVNNs which, unlike PCA, exploit all PCs of the data.
Note that $\Cbal$ in~\eqref{eq::cbal} is only defined for two groups, but extensions to more than two groups are possible~\cite{kamani2022efficient}, while $\Cdeb$ in~\eqref{eq::cdeb} applies to any number of groups $G$.

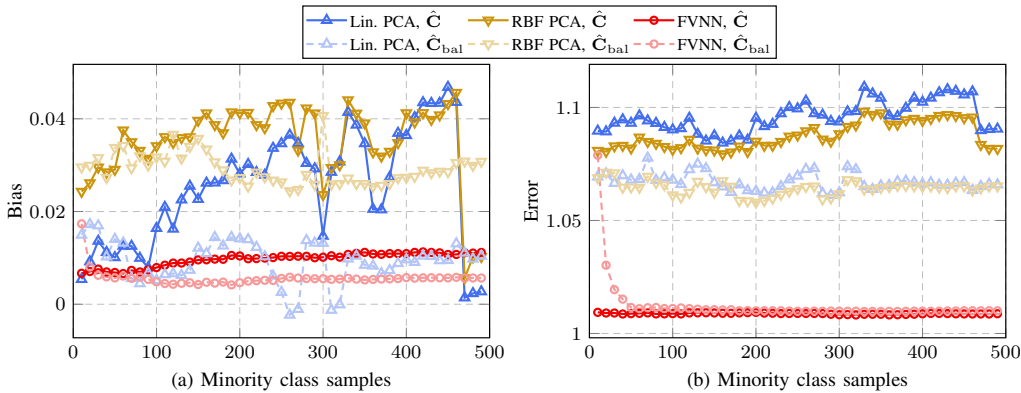
\begin{figure*}[t]
    \centering
    \begin{minipage}[b][][b]{.30\textwidth}
        \begin{tikzpicture}[baseline,scale=.9,trim axis left, trim axis right]

\pgfmathsetmacro{\initmarksize}{2.5}

\pgfplotstableread{data/synth_smape_diff.csv}\errtable

\begin{axis}[
    xlabel={(a) Minority class samples},
    ylabel={Bias},
    xmin=0,
    xmax=500,
    yticklabel style={
        /pgf/number format/fixed,
        /pgf/number format/precision=3
    },
    scaled y ticks=false,
    grid style=densely dashed,
    grid=both,
    legend style={
        at={(.55, 1.02)},
        anchor=south west},
    legend columns=3,
    width=220,
    height=160,
    label style={font=\small},
    tick label style={font=\small}
    ]

    \addplot[ \lpcasmp, 
        mark=triangle, 
        solid, 
        mark size=1.5 ]
    table [ x=min_class_samples, y=PCA_lin ] {\errtable};

    \addplot[ \kpcasmp, 
        mark=triangle, 
        solid, 
        every mark/.append style={rotate=180}, 
        mark size=1.5 ]
    table [ x=min_class_samples, y=PCA_RBF ] {\errtable};

    \addplot[ \fvnnsmp, 
        mark=o,
        solid, 
        mark size=1 ]
    table [ x=min_class_samples, y=VNN ] {\errtable};

    \addplot[ \lpcabal, 
        mark=triangle, 
        densely dashed, 
        mark size=1.5 ]
    table [ x=min_class_samples, y=PCA_lin_fair ] {\errtable};

    \addplot[ \kpcabal, 
        mark=triangle, 
        densely dashed, 
        every mark/.append style={rotate=180}, 
        mark size=1.5 ]
    table [ x=min_class_samples, y=PCA_RBF_fair ] {\errtable};

    \addplot[ \fvnnbal, 
        mark=o,
        densely dashed, 
        mark size=1 ]
    table [ x=min_class_samples, y=VNN_fair ] {\errtable};

    
    \legend{
        {Lin. PCA, $\hbC$},
        {RBF PCA, $\hbC$},
        {FVNN, $\hbC$},
        {Lin. PCA, $\Cbal$},
        {RBF PCA, $\Cbal$},
        {FVNN, $\Cbal$}
    }

\end{axis}

\end{tikzpicture}
    \end{minipage}
    \hspace{1.2cm}
    \begin{minipage}[b][][b]{.30\textwidth}
        \begin{tikzpicture}[baseline,scale=.9,trim axis left, trim axis right]

\pgfmathsetmacro{\initmarksize}{2.5}

\pgfplotstableread{data/synth_smape.csv}\errtable

\begin{axis}[
    xlabel={(b) Minority class samples},
    ylabel={Error},
    xmin=0,
    xmax=500,
    yticklabel style={
        /pgf/number format/fixed,
        /pgf/number format/precision=3
    },
    grid style=densely dashed,
    grid=both,
    legend style={
        at={(.5, 1.02)},
        anchor=south},
    legend columns=3,
    width=220,
    height=160,
    label style={font=\small},
    tick label style={font=\small}
    ]

    \addplot[ \lpcasmp, 
        mark=triangle, 
        solid, 
        mark size=1.5 ]
    table [ x=min_class_samples, y=PCA_lin ] {\errtable};

    \addplot[ \kpcasmp, 
        mark=triangle, 
        solid, 
        every mark/.append style={rotate=180}, 
        mark size=1.5 ]
    table [ x=min_class_samples, y=PCA_RBF ] {\errtable};

    \addplot[ \fvnnsmp, 
        mark=o,
        solid, 
        mark size=1 ]
    table [ x=min_class_samples, y=VNN ] {\errtable};

    \addplot[ \lpcabal, 
        mark=triangle, 
        densely dashed, 
        mark size=1.5 ]
    table [ x=min_class_samples, y=PCA_lin_fair ] {\errtable};

    \addplot[ \kpcabal, 
        mark=triangle, 
        densely dashed, 
        every mark/.append style={rotate=180}, 
        mark size=1.5 ]
    table [ x=min_class_samples, y=PCA_RBF_fair ] {\errtable};

    \addplot[ \fvnnbal, 
        mark=o,
        densely dashed, 
        mark size=1 ]
    table [ x=min_class_samples, y=VNN_fair ] {\errtable};



\end{axis}

\end{tikzpicture}
    \end{minipage}
    \caption{
    Performance of PCA-based and FVNN models for a synthetic regression task.
    Each model is compared using the sample covariance $\hbC$ and the balanced covariance $\Cbal$.
    (a) Fairness measured as imbalance in sMAPE across groups.
    (b) Error measured as sMAPE.
    The legend is shared by both plots.
    }
\label{f:synth_exps}
\end{figure*}

\subsection{Bias mitigation penalties}
\label{ss:bp}

In addition to providing fairer data information, we can also encourage unbiased behavior by manipulating the training loss.
We can formulate the training objective as 
\alna{
    \min_{\ccalH} ~ \gamma \ccalL(\bbX, \bby, \Phi) + (1-\gamma) \ccalR( \bbX, \bby, \bbz, \Phi ),
\label{eq:bias_penalty_generic}
}
where $\Phi$ is the VNN using $\Cbal$ or $\Cdeb$, $\ccalL$ is the task-specific loss function, $\ccalR$ denotes a bias metric, and $\gamma\in[0,1]$ is a balancing weight to tune between $\ccalL$ and $\ccalR$. 
The penalty $\ccalR$ measures group-wise imbalance when using $\bbX$ and $\Phi$ to predict $\bby$.
The choice of $\ccalR$ is entirely flexible, including popular bias metrics DP or EO.
In our case, we let $\ccalR$ be the group-wise imbalance in accuracy $\Delta \ccalL$ in~\eqref{eq:imbgrp}.

\section{Theoretical Analysis}
\label{sec:theory}
Fair covariance estimation approaches such as $\Cbal$ and $\Cdeb$ are notoriously unstable in small data regimes or when the eigenvalues of the covariance matrix are close, that is, finite sample estimation errors may lead to significant differences in the estimated fair covariance matrix and, consequently, in their PCs~\cite{Jolliffe2002pca}. 
This can lead to unfair behavior when samples from different groups are unbalanced or distributions differ across groups.
In this context, the stability of VNNs to covariance estimation errors intrinsically improves fairness, as we discuss in the following theorem. 

\begin{mydefinition}\label{def:lipschitz}
Let $h(\lambda)$ be the frequency response of the filter $\bbH(\bbC)$, which evaluates the behavior of the filter in the spectral domain at eigenvalues $\lambda$ of the covariance matrix~\cite{sihag2022covariance}.
The filter $\bbH(\bbC)$ is Lipschitz with constant $P$ if $|h(\lambda_i) - h(\lambda_j)| \leq P|\lambda_i - \lambda_j| $ for every eigenvalue pair $\lambda_i, \lambda_j, i \neq j$.
\end{mydefinition}

\begin{mytheorem}
\label{th:stability_fairness}
    Consider a covariance filter $\bbH(\bbC)$ that is Lipschitz with constant $P$ as per Def.~\ref{def:lipschitz}. 
    
    First, consider two groups with true covariances $\bbC_g, \bbC_h$. We consider the fair covariance estimate $\Cbal$ from~\eqref{eq::cbal} with $T$ samples and we assume the fair true covariance to be $\bbC = \alpha_g \bbC_g + \alpha_h \bbC_h$ with $\alpha_g,\alpha_h$ in~\eqref{eq::cbal}. We express VNN stability as
    \begin{align}
        \| \bbH(\bbC) - \bbH(\Cbal) \| \leq P\sqrt{N+2N^2} \left( \mathcal{O}(T_g^{-1/2}) + \mathcal{O}(T_h^{-1/2}) \right).
    \end{align}

    Second, let $\bbC$ be any covariance matrix, but the observed data $\bbX$ is biased with covariance $\mbE[\bbX^\top (\bbI_T + \beta \bbZ\bbZ^\top) \bbX]$, where we let the mean $\bbmu$ be zero for simplicity.
    Then, given the fair covariance estimate $\Cdeb$ in~\eqref{eq::cdeb} using $T$ samples, we write VNN stability as
    \begin{align}
        \| \bbH(\bbC) - \bbH(\Cdeb) \| \leq P\sqrt{N+2N^2} \mathcal{O}(T^{-1/2}).
    \end{align}
\end{mytheorem}
\begin{myproof}
    Following~\cite[Appendix C, Proposition 2]{cavallo2024sparsecovarianceneuralnetworks}, for two generic true and sample covariances $\bbC, \hbC$, we can write 
    \begin{align}
    \label{eq:vnn_stab_generic}
        \| \bbH(\bbC) - \bbH(\hat{\bbC}) \| \leq P\sqrt{N+2N^2}\|\bbE\| + \mathcal{O}(\|\bbE\|^2),
    \end{align}
    where $\bbE = \bbC - \hat{\bbC}$ and $\|\bbE\|^2$ is negligible for $T$ large enough. 
    
    For the first case, by the triangle inequality, we have
    \begin{align}
        \| \Cbal - \bbC \| \leq |\alpha_h| \| \hat{\bbC}_h - \bbC_h\| + |\alpha_g|\| \hat{\bbC}_g - \bbC_g  \|,
    \end{align}
    where each term in the norm is the estimation error of a covariance matrix, which decreases as the inverse square root of the number of samples, that is, $\| \Cbal - \bbC \| \leq \mathcal{O}(T_g^{-1/2}) + \mathcal{O}(T_h^{-1/2})$ with high probability~\cite[Theorem 5.6.1]{vershynin2018high}. Replacing this in~\eqref{eq:vnn_stab_generic}, we obtain the bound for $\Cbal$.
    
    For the second case, observe that the true covariance matrix $\bbC$ generates the unbiased samples $\bbX' = (\bbI_T + \beta \bbZ\bbZ^\top)^{-1/2}\bbX$, which are the transformed samples used to estimate $\Cdeb$.
    In this case, $\Cdeb$ is the classic sample covariance estimator for the samples $\bbX'$, so we have that $\|\Cdeb - \bbC\| \leq \mathcal{O}(T^{-1/2})$~\cite[Theorem 5.6.1]{vershynin2018high}. Combining this with~\eqref{eq:vnn_stab_generic} leads to the bound for $\Cdeb$.
\end{myproof}

Theorem~\ref{th:stability_fairness} shows that a covariance filter $\bbH(\bbC)$ and thus a VNN~\cite[Theorem 3]{sihag2022covariance} operating on a fair sample covariance estimate is stable to estimation errors.
In particular, we may design the filter $\bbH(\bbC)$ while considering the Lipschitz constant $P$, allowing us to control the tradeoff between stability and discriminability.
On the contrary, PCA does not have this flexibility since its stability depends inversely on the smallest gap in covariance eigenvalues~\cite[Proposition 1]{cavallo2024stvnn}.
Covariance estimation error may differ across groups if one group has fewer samples, but Theorem~\ref{th:stability_fairness} shows that FVNNs achieve a more consistent behavior across groups compared to PCA and therefore intrinsically provide superior fairness.

\section{Numerical evaluation}
\label{s:results}

\subsection{Synthetic data}

\begin{figure*}[t]
    \centering
    \hspace{0.7cm}
    \begin{minipage}[b][][b]{.30\textwidth}
        \begin{tikzpicture}[baseline,scale=.8,trim axis left, trim axis right]

\pgfmathsetmacro{\initmarksize}{2.5}

\pgfplotstableread{data/vnn_parkinson.csv}\vnntable
\pgfplotstableread{data/pca_parkinson.csv}\pcatable

\begin{axis}[
    xlabel={Error},
    ylabel={Bias},
    yticklabel style={
        /pgf/number format/fixed,
        /pgf/number format/precision=3
    },
    grid style=densely dashed,
    grid=both,
    legend style={
        at={(.5, 1.02)},
        anchor=south},
    legend columns=3,
    width=220,
    height=170,
    label style={font=\small},
    tick label style={font=\small}
    ]
    
    \addplot[ \lpcasmp, 
        mark=triangle, mark size=2, 
        solid, 
        error bars/.cd, y dir=both, y explicit, x dir=both, x explicit, error mark=none]  
        table [ x=PCA_lin_acc, y=PCA_lin_fair, y error=PCA_lin_fair_std, x error=PCA_lin_acc_std ] {\pcatable};
    \addplot[ mark options={\lpcasmp}, mark repeat=8, 
        mark size=2, mark=triangle*, 
        only marks, forget plot ] 
        table [ x=PCA_lin_acc, y=PCA_lin_fair ] {\pcatable};

    \addplot[ \lpcadeb, 
        mark=triangle, mark size=2, 
        densely dotted, 
        error bars/.cd, y dir=both, y explicit, x dir=both, x explicit, error mark=none]  
        table [ x=PCA_lin_deb_acc, y=PCA_lin_deb_fair, y error=PCA_lin_deb_fair_std, x error=PCA_lin_deb_acc_std ] {\pcatable};
    \addplot[ mark options={\lpcadeb}, mark repeat=8, 
        mark size=2, mark=triangle*, 
        only marks, forget plot ] 
        table [ x=PCA_lin_deb_acc, y=PCA_lin_deb_fair ] {\pcatable};
        
    \addplot[ \lpcabal, 
        mark=triangle, mark size=2, 
        densely dashed, 
        error bars/.cd, y dir=both, y explicit, x dir=both, x explicit, error mark=none]  
        table [ x=PCA_lin_bal_acc, y=PCA_lin_bal_fair, y error=PCA_lin_bal_fair_std, x error=PCA_lin_bal_acc_std ] {\pcatable};
    \addplot[ mark options={\lpcabal}, mark repeat=8, 
        mark size=2, mark=triangle*, 
        only marks, forget plot ] 
        table [ x=PCA_lin_bal_acc, y=PCA_lin_bal_fair ] {\pcatable};

    \addplot[ \kpcasmp, 
        mark=triangle, mark size=2, 
        every mark/.append style={rotate=180}, 
        solid, 
        error bars/.cd, y dir=both, y explicit, x dir=both, x explicit, error mark=none]  
        table [ x=PCA_RBF_acc, y=PCA_RBF_fair, y error=PCA_RBF_fair_std, x error=PCA_RBF_acc_std ] {\pcatable};
    \addplot[ mark options={\kpcasmp}, mark repeat=8, 
        mark size=2, mark=triangle*, 
        every mark/.append style={rotate=180}, 
        only marks, forget plot ] 
        table [ x=PCA_RBF_acc, y=PCA_RBF_fair ] {\pcatable};

    \addplot[ \kpcadeb, 
        mark=triangle, mark size=2, 
        every mark/.append style={rotate=180}, 
        densely dotted, 
        error bars/.cd, y dir=both, y explicit, x dir=both, x explicit, error mark=none]  
        table [ x=PCA_RBF_deb_acc, y=PCA_RBF_deb_fair, y error=PCA_RBF_deb_fair_std, x error=PCA_RBF_deb_acc_std ] {\pcatable};
    \addplot[ mark options={\kpcadeb}, mark repeat=8, 
        mark size=2, mark=triangle*, 
        every mark/.append style={rotate=180}, 
        only marks, forget plot ] 
        table [ x=PCA_RBF_deb_acc, y=PCA_RBF_deb_fair ] {\pcatable};
        
    \addplot[ \kpcabal, 
        mark=triangle, mark size=2, 
        every mark/.append style={rotate=180}, 
        densely dashed, 
        error bars/.cd, y dir=both, y explicit, x dir=both, x explicit, error mark=none]  
        table [ x=PCA_RBF_bal_acc, y=PCA_RBF_bal_fair, y error=PCA_RBF_bal_fair_std, x error=PCA_RBF_bal_acc_std ] {\pcatable};
    \addplot[ mark options={\kpcabal}, mark repeat=8, 
        mark size=2, mark=triangle*, 
        every mark/.append style={rotate=180}, 
        only marks, forget plot ] 
        table [ x=PCA_RBF_bal_acc, y=PCA_RBF_bal_fair ] {\pcatable};

    \addplot[ \fvnnsmp, 
        mark=o, mark size=1.5, 
        solid, 
        error bars/.cd, y dir=both, y explicit, x dir=both, x explicit, error mark=none]  
        table [ x=VNN_acc, y=VNN_fair, y error=VNN_fair_std, x error=VNN_acc_std ] {\vnntable};
    \addplot[ mark options={\fvnnsmp}, mark repeat=8, 
        mark size=1.5, mark=*, 
        only marks, forget plot ] 
        table [ x=VNN_acc, y=VNN_fair ] {\vnntable};

    \addplot[ \fvnndeb, 
        mark=o, mark size=1.5, 
        densely dotted, 
        error bars/.cd, y dir=both, y explicit, x dir=both, x explicit, error mark=none]  
        table [ x=VNN_deb_acc, y=VNN_deb_fair, y error=VNN_deb_fair_std, x error=VNN_deb_acc_std ] {\vnntable};
    \addplot[ mark options={\fvnndeb}, mark repeat=8, mark size=1.5, only marks, mark=*, forget plot ] 
        table [ x=VNN_deb_acc, y=VNN_deb_fair ] {\vnntable};
        
    \addplot[ \fvnnbal, 
        mark=o, mark size=1.5, 
        densely dashed, 
        error bars/.cd, y dir=both, y explicit, x dir=both, x explicit, error mark=none]  
        table [ x=VNN_bal_acc, y=VNN_bal_fair, y error=VNN_bal_fair_std, x error=VNN_bal_acc_std ] {\vnntable};
    \addplot[ mark options={\fvnnbal}, mark repeat=8, 
        mark size=1.5, 
        only marks, mark=*, forget plot ] 
        table [ x=VNN_bal_acc, y=VNN_bal_fair ] {\vnntable};
        

\end{axis}

\end{tikzpicture}
    \end{minipage}
    \hspace{.3cm}
    \begin{minipage}[b][][b]{.30\textwidth}
        \begin{tikzpicture}[baseline,scale=.8,trim axis left, trim axis right]

\pgfmathsetmacro{\initmarksize}{2.5}

\pgfplotstableread{data/vnn_lsac.csv}\vnntable
\pgfplotstableread{data/pca_lsac.csv}\pcatable

\begin{axis}[
    xlabel={Error},
    ylabel={Bias},
    yticklabel style={
        /pgf/number format/fixed,
        /pgf/number format/precision=3
    },
    grid style=densely dashed,
    grid=both,
    legend style={
        at={(.5, 1.02)},
        anchor=south},
    legend columns=3,
    width=220,
    height=170,
    label style={font=\small},
    tick label style={font=\small}
    ]
        
    \addplot[ \lpcasmp, 
        mark=triangle, mark size=2, 
        solid, 
        error bars/.cd, y dir=both, y explicit, x dir=both, x explicit, error mark=none]  
        table [ x=PCA_lin_acc, y=PCA_lin_fair, y error=PCA_lin_fair_std, x error=PCA_lin_acc_std ] {\pcatable};
    \addplot[ mark options={\lpcasmp}, mark repeat=8, 
        mark size=2, mark=triangle*, 
        only marks, forget plot ] 
        table [ x=PCA_lin_acc, y=PCA_lin_fair ] {\pcatable};

    \addplot[ \lpcadeb, 
        mark=triangle, mark size=2, 
        densely dotted, 
        error bars/.cd, y dir=both, y explicit, x dir=both, x explicit, error mark=none]  
        table [ x=PCA_lin_deb_acc, y=PCA_lin_deb_fair, y error=PCA_lin_deb_fair_std, x error=PCA_lin_deb_acc_std ] {\pcatable};
    \addplot[ mark options={\lpcadeb}, mark repeat=8, 
        mark size=2, mark=triangle*, 
        only marks, forget plot ] 
        table [ x=PCA_lin_deb_acc, y=PCA_lin_deb_fair ] {\pcatable};
        
    \addplot[ \lpcabal, 
        mark=triangle, mark size=2, 
        densely dashed, 
        error bars/.cd, y dir=both, y explicit, x dir=both, x explicit, error mark=none]  
        table [ x=PCA_lin_bal_acc, y=PCA_lin_bal_fair, y error=PCA_lin_bal_fair_std, x error=PCA_lin_bal_acc_std ] {\pcatable};
    \addplot[ mark options={\lpcabal}, mark repeat=8, 
        mark size=2, mark=triangle*, 
        only marks, forget plot ] 
        table [ x=PCA_lin_bal_acc, y=PCA_lin_bal_fair ] {\pcatable};

    \addplot[ \kpcasmp, 
        mark=triangle, mark size=2, 
        every mark/.append style={rotate=180}, 
        solid, 
        error bars/.cd, y dir=both, y explicit, x dir=both, x explicit, error mark=none]  
        table [ x=PCA_RBF_acc, y=PCA_RBF_fair, y error=PCA_RBF_fair_std, x error=PCA_RBF_acc_std ] {\pcatable};
    \addplot[ mark options={\kpcasmp}, mark repeat=8, 
        mark size=2, mark=triangle*, 
        every mark/.append style={rotate=180}, 
        only marks, forget plot ] 
        table [ x=PCA_RBF_acc, y=PCA_RBF_fair ] {\pcatable};

    \addplot[ \kpcadeb, 
        mark=triangle, mark size=2, 
        every mark/.append style={rotate=180}, 
        densely dotted, 
        error bars/.cd, y dir=both, y explicit, x dir=both, x explicit, error mark=none]  
        table [ x=PCA_RBF_deb_acc, y=PCA_RBF_deb_fair, y error=PCA_RBF_deb_fair_std, x error=PCA_RBF_deb_acc_std ] {\pcatable};
    \addplot[ mark options={\kpcadeb}, mark repeat=8, 
        mark size=2, mark=triangle*, 
        every mark/.append style={rotate=180}, 
        only marks, forget plot ] 
        table [ x=PCA_RBF_deb_acc, y=PCA_RBF_deb_fair ] {\pcatable};
        
    \addplot[ \kpcabal, 
        mark=triangle, mark size=2, 
        every mark/.append style={rotate=180}, 
        densely dashed, 
        error bars/.cd, y dir=both, y explicit, x dir=both, x explicit, error mark=none]  
        table [ x=PCA_RBF_bal_acc, y=PCA_RBF_bal_fair, y error=PCA_RBF_bal_fair_std, x error=PCA_RBF_bal_acc_std ] {\pcatable};
    \addplot[ mark options={\kpcabal}, mark repeat=8, 
        mark size=2, mark=triangle*, 
        every mark/.append style={rotate=180}, 
        only marks, forget plot ] 
        table [ x=PCA_RBF_bal_acc, y=PCA_RBF_bal_fair ] {\pcatable};

    \addplot[ \fvnnsmp, 
        mark=o, mark size=1.5, 
        solid, 
        error bars/.cd, y dir=both, y explicit, x dir=both, x explicit, error mark=none]  
        table [ x=VNN_acc, y=VNN_fair, y error=VNN_fair_std, x error=VNN_acc_std ] {\vnntable};
    \addplot[ mark options={\fvnnsmp}, mark repeat=8, 
        mark size=1.5, mark=*, 
        only marks, forget plot ] 
        table [ x=VNN_acc, y=VNN_fair ] {\vnntable};

    \addplot[ \fvnndeb, 
        mark=o, mark size=1.5, 
        densely dotted, 
        error bars/.cd, y dir=both, y explicit, x dir=both, x explicit, error mark=none]  
        table [ x=VNN_deb_acc, y=VNN_deb_fair, y error=VNN_deb_fair_std, x error=VNN_deb_acc_std ] {\vnntable};
    \addplot[ mark options={\fvnndeb}, mark repeat=8, mark size=1.5, only marks, mark=*, forget plot ] 
        table [ x=VNN_deb_acc, y=VNN_deb_fair ] {\vnntable};
        
    \addplot[ \fvnnbal, 
        mark=o, mark size=1.5, 
        densely dashed, 
        error bars/.cd, y dir=both, y explicit, x dir=both, x explicit, error mark=none]  
        table [ x=VNN_bal_acc, y=VNN_bal_fair, y error=VNN_bal_fair_std, x error=VNN_bal_acc_std ] {\vnntable};
    \addplot[ mark options={\fvnnbal}, mark repeat=8, 
        mark size=1.5, 
        only marks, mark=*, forget plot ] 
        table [ x=VNN_bal_acc, y=VNN_bal_fair ] {\vnntable};
        
    \legend{
        {Lin. PCA, $\hbC$},
        {Lin. PCA, $\Cdeb$},
        {Lin. PCA, $\Cbal$},
        {RBF PCA, $\hbC$},
        {RBF PCA, $\Cdeb$},
        {RBF PCA, $\Cbal$},
        {FVNN, $\hbC$},
        {FVNN, $\Cdeb$},
        {FVNN, $\Cbal$}
    }

\end{axis}

\end{tikzpicture}
    \end{minipage}
    \hspace{.3cm}
    \begin{minipage}[b][][b]{.30\textwidth}
        \begin{tikzpicture}[baseline,scale=.8,trim axis left, trim axis right]

\pgfmathsetmacro{\initmarksize}{2.5}

\pgfplotstableread{data/credit_vnn_flip.csv}\vnntable
\pgfplotstableread{data/credit_pca_flip.csv}\pcatable

\begin{axis}[
    xlabel={Error},
    ylabel={Bias},
    yticklabel style={
        /pgf/number format/fixed,
        /pgf/number format/precision=3
    },
    grid style=densely dashed,
    grid=both,
    legend style={
        at={(.5, 1.02)},
        anchor=south},
    legend columns=3,
    width=220,
    height=170,
    label style={font=\small},
    tick label style={font=\small}
    ]

    \addplot[ \kpcasmp, 
        mark=triangle, mark size=2, 
        every mark/.append style={rotate=180}, 
        solid, 
        error bars/.cd, y dir=both, y explicit, x dir=both, x explicit, error mark=none]  
        table [ x=pcaSampRbf-acc, y=pcaSampRbf-fair, y error=pcaSampRbf-fair-std, x error=pcaSampRbf-acc-std ] {\pcatable};
    \addplot[ mark options={\kpcasmp}, mark repeat=8, 
        mark size=2, mark=triangle*, 
        every mark/.append style={rotate=180}, 
        only marks, forget plot ] 
        table [ x=pcaSampRbf-acc, y=pcaSampRbf-fair ] {\pcatable};

    \addplot[ \kpcadeb, 
        mark=triangle, mark size=2, 
        every mark/.append style={rotate=180}, 
        densely dotted, 
        error bars/.cd, y dir=both, y explicit, x dir=both, x explicit, error mark=none]  
        table [ x=pcaDebRbf-acc, y=pcaDebRbf-fair, y error=pcaDebRbf-fair-std, x error=pcaDebRbf-acc-std ] {\pcatable};
    \addplot[ mark options={\kpcadeb}, mark repeat=8, 
        mark size=2, mark=triangle*, 
        every mark/.append style={rotate=180}, 
        only marks, forget plot ] 
        table [ x=pcaDebRbf-acc, y=pcaDebRbf-fair ] {\pcatable};
        
    \addplot[ \kpcabal, 
        mark=triangle, mark size=2, 
        every mark/.append style={rotate=180}, 
        densely dashed, 
        error bars/.cd, y dir=both, y explicit, x dir=both, x explicit, error mark=none]  
        table [ x=pcaBalRbf-acc, y=pcaBalRbf-fair, y error=pcaBalRbf-fair-std, x error=pcaBalRbf-acc-std ] {\pcatable};
    \addplot[ mark options={\kpcabal}, mark repeat=8, 
        mark size=2, mark=triangle*, 
        every mark/.append style={rotate=180}, 
        only marks, forget plot ] 
        table [ x=pcaBalRbf-acc, y=pcaBalRbf-fair ] {\pcatable};
        
    \addplot[ \lpcasmp, 
        mark=triangle, mark size=2, 
        solid, 
        error bars/.cd, y dir=both, y explicit, x dir=both, x explicit, error mark=none]  
        table [ x=pcaSampLin-acc, y=pcaSampLin-fair, y error=pcaSampLin-fair-std, x error=pcaSampLin-acc-std ] {\pcatable};
    \addplot[ mark options={\lpcasmp}, mark repeat=8, 
        mark size=2, mark=triangle*, 
        only marks, forget plot ] 
        table [ x=pcaSampLin-acc, y=pcaSampLin-fair ] {\pcatable};

    \addplot[ \lpcadeb, 
        mark=triangle, mark size=2, 
        densely dotted, 
        error bars/.cd, y dir=both, y explicit, x dir=both, x explicit, error mark=none]  
        table [ x=pcaDebLin-acc, y=pcaDebLin-fair, y error=pcaDebLin-fair-std, x error=pcaDebLin-acc-std ] {\pcatable};
    \addplot[ mark options={\lpcadeb}, mark repeat=8, 
        mark size=2, mark=triangle*, 
        only marks, forget plot ] 
        table [ x=pcaDebLin-acc, y=pcaDebLin-fair ] {\pcatable};
        
    \addplot[ \lpcabal, 
        mark=triangle, mark size=2, 
        densely dashed, 
        error bars/.cd, y dir=both, y explicit, x dir=both, x explicit, error mark=none]  
        table [ x=pcaBalLin-acc, y=pcaBalLin-fair, y error=pcaBalLin-fair-std, x error=pcaBalLin-acc-std ] {\pcatable};
    \addplot[ mark options={\lpcabal}, mark repeat=8, 
        mark size=2, mark=triangle*, 
        only marks, forget plot ] 
        table [ x=pcaBalLin-acc, y=pcaBalLin-fair ] {\pcatable};

    \addplot[ \fvnnsmp, 
        mark=o, mark size=1.5, 
        solid, 
        error bars/.cd, y dir=both, y explicit, x dir=both, x explicit, error mark=none]  
        table [ x=vnnSamp-acc, y=vnnSamp-fair, y error=vnnSamp-fair-std, x error=vnnSamp-acc-std ] {\vnntable};
    \addplot[ mark options={\fvnnsmp}, mark repeat=8, 
        mark size=1.5, mark=*, 
        only marks, forget plot ] 
        table [ x=vnnSamp-acc, y=vnnSamp-fair ] {\vnntable};

    \addplot[ \fvnndeb,
        mark=o, mark size=1.5, 
        densely dotted, 
        error bars/.cd, y dir=both, y explicit, x dir=both, x explicit, error mark=none]  
        table [ x=vnnDeb-acc, y=vnnDeb-fair, y error=vnnDeb-fair-std, x error=vnnDeb-acc-std ] {\vnntable};
    \addplot[ mark options={\fvnndeb}, mark repeat=8, 
        mark size=1.5, mark=*, 
        only marks, forget plot ] 
        table [ x=vnnDeb-acc, y=vnnDeb-fair ] {\vnntable};
        
    \addplot[ \fvnnbal,
        mark=o, mark size=1.5, 
        densely dashed, 
        error bars/.cd, y dir=both, y explicit, x dir=both, x explicit, error mark=none]  
        table [ x=vnnBal-acc, y=vnnBal-fair, y error=vnnBal-fair-std, x error=vnnBal-acc-std ] {\vnntable};
    \addplot[ mark options={\fvnnbal}, mark repeat=8, 
        mark size=1.5, mark=*, 
        only marks, forget plot ] 
        table [ x=vnnBal-acc, y=vnnBal-fair ] {\vnntable};


\end{axis}

\end{tikzpicture}
    \end{minipage}
    \caption{
    Performance of PCA-based and FVNN models for real-world regression and classification tasks.
    For each plot, the $y$-axis denotes bias and the $x$-axis error. 
    (a) Parkinson regression as the bias penalty weight $\gamma$ increases. Results for PCA with $\Cdeb$ are overlapped with those with $\hbC$, which are therefore not visible.
    (b) LSAC regression as the bias penalty weight $\gamma$ increases. 
    (c) German Credit classification. \textbf{FVNN} is shown with and without a bias penalty, while PCA-based models are shown with 10 and 30 PCs.
    The legend is shared by all three plots.
    }
\label{f:real_exps}
\end{figure*}
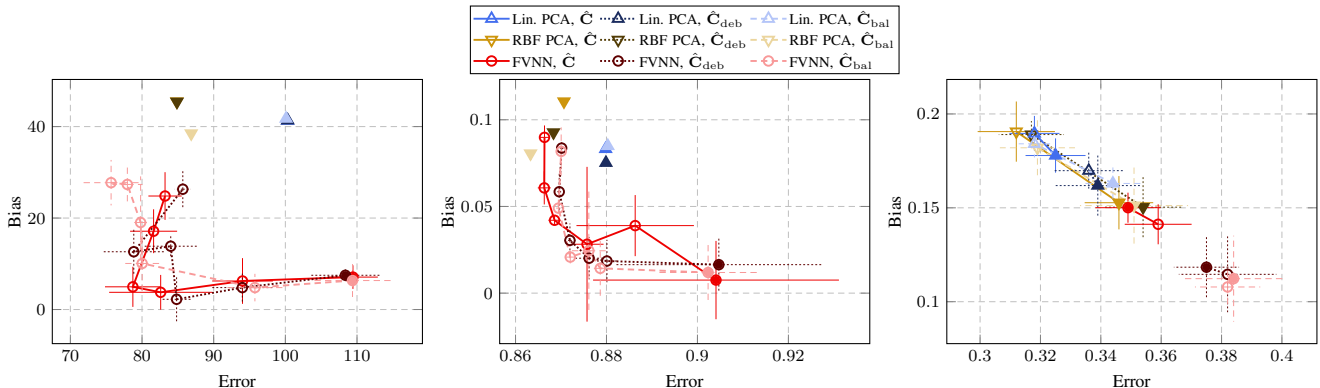

\noindent \textbf{Experimental setup.} 
To validate the impact of FVNN stability on fairness, we generate a synthetic dataset with $G=2$ groups, where features are sampled corresponding to two different multivariate Gaussian distributions with different covariance matrices $\bbC_1$ and $\bbC_2$.
We generate a regression target following the Friedman regression problem~\cite{friedman1991}.
We let group 1 be the disadvantaged group, that is, the eigenvalues for $\bbC_1$ are closer than those of $\bbC_2$, rendering the estimation of $\bbC_1$ more difficult.
While the training data is balanced between groups, in testing, we replace the sample covariance matrix used in training with an estimate obtained from testing data as the number of samples $T_1$ in group 1 increases from $1$ to $500$, while we fix $T_2=500$. We compute the test performance on the entire test set, that is, with $T_1=T_2=500$.
We compare FVNN performance, denoted ``\textbf{FVNN}'', to SVM regression using PCA-projected features, where we apply both linear SVM, denoted ``\textbf{Lin. PCA}'', and kernelized SVM, denoted ``\textbf{RBF PCA}''.
Moreover, for each method we apply the original sample covariance $\hbC$ and the balanced covariance estimate $\Cbal$ with $\alpha = 0.5$.
We measure error $\ccalL$ as the symmetric mean average percentage error (sMAPE), while the bias is measured as $\Delta\ccalL$ as in~\eqref{eq:imbgrp}. 
\medskip

\noindent \textbf{Discussion.} 
Fig.~\ref{f:synth_exps} shows that VNNs are significantly more stable both in terms of fairness and prediction performance compared to PCA-based variants, which are more susceptible to changes in the covariance matrix. 
Indeed, even a small number of new samples can yield significant changes in bias and error for \textbf{Lin. PCA} and \textbf{RBF PCA}, while \textbf{FVNN} returns smooth estimates as the sample ratio between groups varies.
Thus, VNNs are more reliable for fair learning under noisy covariance estimates, particularly when one group is more difficult to estimate. 
Moreover, we not only observe improved bias in Fig.~\ref{f:synth_exps}a when using $\Cbal$ in place of $\hbC$, but VNNs also outperform PCA-based models in terms of error with either covariance matrix estimate in Fig.~\ref{f:synth_exps}b.

\subsection{Real data - regression}

\noindent \textbf{Experimental setup.} 
Next, we apply FVNNs for regression tasks using two real datasets with known biases.
Parkinson~\cite{tsanas2009accurate} contains 5,875 records of 23 features for 42 patients with early-stage Parkinson's disease. 
The objective is to predict, for each record, the \textit{Unified Parkinson's Disease Rating Scale score}, a continuous value measuring different aspects of Parkinson's disease. 
The sensitive attribute is the sex of the patient (female 33\%, male 67\%).
Law School Admission Council (LSAC)~\cite{Wightman1998LSACNL} contains 5 features for 22,407 law school students (22,368 without missing data). 
The target is the \textit{Grade Point Average} and we use as sensitive attribute the race of students (white/Caucasian 88.2\%, other 11.8\%).


We again compare \textbf{FVNN} to \textbf{Lin. PCA} and \textbf{RBF PCA}, which respectively use linear and kernelized SVM for regression.
For all three methods, we compare the sample covariance $\hbC$ with the fair estimates $\Cbal$ and $\Cdeb$, where we select $\alpha$ and $\beta$ through a grid search, along with the VNN size and number of PCs.
Furthermore, we employ the bias penalty $\ccalR$ as in~\eqref{eq:imbgrp} for the training loss in~\eqref{eq:bias_penalty_generic}, where $\ccalL$ denotes the mean squared error (MSE).
We vary the penalty weight from $\gamma=0.3$ (filled markers) to $\gamma=1$.
Figs.~\ref{f:real_exps}a and b show the average results and standard deviation over 5 trials.

\medskip

\noindent \textbf{Discussion.}
We observe that \textbf{FVNN} provides a significantly more flexible tool to control the fairness-accuracy tradeoff compared to the PCA-based models. 
Smaller values of $\gamma$ lead to fairer solutions for \textbf{FVNN} at the expense of a higher regression error.
For large enough $\gamma$, \textbf{FVNN} with any covariance matrix outperforms \textbf{Lin. PCA} and \textbf{RBF PCA} in both fairness and accuracy.
Applying $\Cbal$ improves bias for \textbf{RBF PCA} but not \textbf{Lin. PCA}, while $\Cdeb$ can decrease bias for both PCA methods on LSAC.
This shows that fair covariance estimators may not lead to improved fairness for downstream tasks, calling for more flexible and powerful solutions.
For our \textbf{FVNN} approach, the fair covariance matrices yield minor improvements in bias for LSAC and negligible differences for Parkinson. 
Thus, we show that real-world datasets may contain biases that cannot be reduced by data preprocessing, but FVNNs offer an effective approach.

\subsection{Real data - classification}

\noindent \textbf{Experimental setup.} 
Finally, we consider a classification task using a real-world German Credit dataset of 1000 individuals.
The goal is to predict credit score as either good or bad for each individual given a set of 46 features, with sex (female 31\%, male 69\%) as the sensitive attribute.
We compare \textbf{FVNN}, \textbf{Lin. PCA}, and \textbf{RBF PCA} for $\hbC$, $\Cdeb$, and $\Cbal$ over 4 train-test splits.
We again consider linear and kernelized SVM for classification.
Fig.~\ref{f:real_exps}c shows the average performance in terms of error, that is, one minus accuracy, and imbalance in error between groups.
For the bias penalty weight, we consider $\gamma=1$, that is, with no penalty $\ccalR$, and $\gamma=0.25$, and we also show the number of PCs as 10 and 30, where the markers for $\gamma=0$ and 10 PCs are filled in Fig.~\ref{f:real_exps}.

\medskip
\noindent \textbf{Discussion.} 
All PCA-based methods increase bias while reducing error when more PCs are considered.
In contrast, when we increase the influence of $\ccalR$ by decreasing $\gamma=1$ to $\gamma=0.25$, \textbf{FVNN} improves the bias at the cost of increased error. 
Importantly, while we see less effect due to $\ccalR$, we note that applying either fair covariance estimate $\Cbal$ or $\Cdeb$ yields greater improvements on bias compared to regression.
This shows that biases in real-world data may not be reduced by a single method, but our proposed \textbf{FVNN} model provides the flexibility to address multiple kinds of bias such as imbalanced representations or unfair outcomes.

\section{Conclusion}
\label{s:conclusion}

In this work, we proposed Fair coVariance Neural Networks (FVNNs), a fairness-aware graph convolutional neural network that operates on the covariance matrix of the data.
FVNNs promote fairness in two ways: by employing a fair covariance matrix to remove biases in data before training and by adding a regularization term in the loss to penalize unfair performance across groups. 
We theoretically showed that FVNNs are intrinsically fairer than fair PCA techniques by building a connection between VNN stability and VNN performance on groups with different distributions.
Furthermore, we empirically validated the efficiency of FVNNs in managing the tradeoff between prediction performance and fairness for multiple applications, showing them to be a significantly more flexible approach than fair PCA.
In future work, we will expand on the effects of biased data and fair interventions for VNN performance beyond stability.
Furthermore, we will address additional notions of fairness beyond balancing performance, such as DP and EO.

\bibliographystyle{ieeetr}
\bibliography{citations}

\end{document}